\documentclass{bmvc2k}

\usepackage{amssymb}
\usepackage{xcolor}
\usepackage{capt-of}
\usepackage{enumitem}
\usepackage{graphicx}
\usepackage{multirow}

\title{Background-Free Objectness Learning for Class-Agnostic Detection}

\addauthor{Dania Batool}{dania.batool@unipa.it}{1}
\addauthor{Liliana Lo Presti}{liliana.lopresti@unipa.it}{1}
\addauthor{Marco La Cascia}{marco.lacascia@unipa.it}{1}
\addauthor{Filippo Vella}{filippo.vella@icar.cnr.it}{2}

\addinstitution{
 Department of Engineering\\
 University of Palermo\\
 Palermo, Italy
}
\addinstitution{
 ICAR\\
 National Research Council of Italy\\
 Palermo, Italy
}

\runninghead{Dania Batool, Liliana Lo Presti, ET AL.}{B-FOR}

\def\etal{\emph{et al}\bmvaOneDot}

\begin{document}
\maketitle
\begin{abstract} 
Object detectors are typically trained under closed-set supervision, where unlabeled regions are implicitly treated as background. Under incomplete annotations, this assumption introduces \emph{objectness bias}: visually valid but unlabeled objects are used as negatives, tying objectness to the annotated taxonomy rather than generic object structure. This limitation is particularly problematic for class-agnostic and open-world detection. This paper proposes \emph{Background-Free Objectness Learning} (B-FOR), a dense class-agnostic detection framework that learns objectness without explicit background supervision on unlabeled regions. B-FOR formulates detection as the prediction of dense multi-scale object-center and scale fields, from which object hypotheses emerge as local spatial structures. Supervision is confined to reliable annotated regions through spatially structured soft targets, avoiding foreground-background discrimination. To support decoding from emergent local maxima, the paper further introduces displacement-aware scale fields that model object extent as a spatially varying property of the learned objectness field. Experiments on PASCAL VOC, MS-COCO, and Open Images demonstrate strong generalization to unseen categories and cross-dataset object distributions. B-FOR improves recall by more than $+10$ AR points over prior class-agnostic baselines. Ablation studies show that both localized objectness supervision and displacement-aware scale fields are critical for class-agnostic localization under incomplete annotations. Code available at: https://github.com/Daniaawan/B-FOR.
\end{abstract}

\section{Introduction}
\label{sec:intro}
\begin{figure*}
\centering
\includegraphics[width=0.85\linewidth]{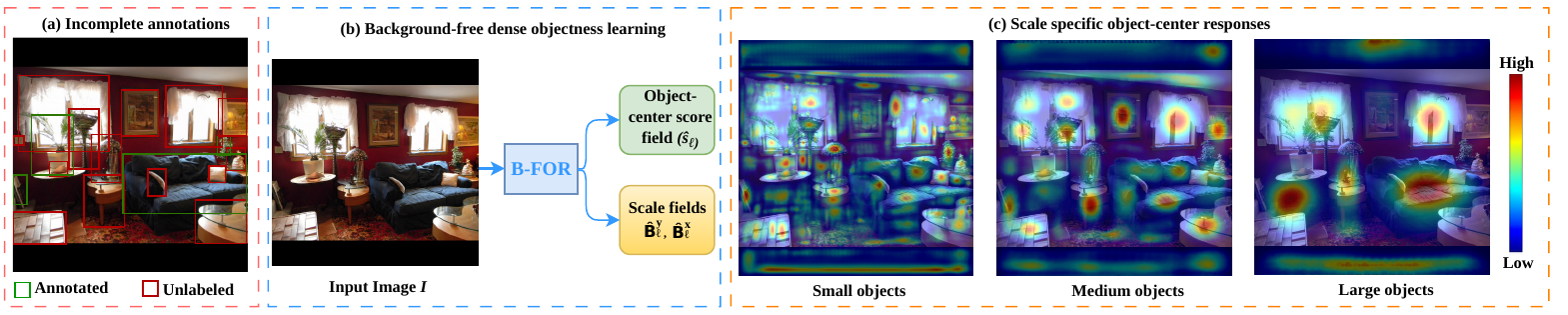}
    \caption{(a) Detection datasets are often incompletely annotated: green boxes indicate labeled objects; red boxes indicate unlabeled object-like regions. (b) B-FOR predicts dense multi-scale object-center and scale fields. (c) Predicted responses consistently activate on both annotated and unlabeled object-like regions across small, medium, and large scales.}
\label{fig:onecol}
\end{figure*}

Object detection has achieved remarkable progress in closed-set settings~\cite{ren2016faster,lin2017focal,tian2019fcos,zhou2019objects,zhang2022dino}, enabled by large-scale benchmarks such as COCO, LVIS, and PASCAL VOC~\cite{lin2014microsoft,gupta2019lvis,everingham2010pascal}. Object localization and representation also underpin related vision tasks such as visual tracking and object matching~\cite{cruciata2021use,presti2008real,monteleone2019particle,lo2009object}. Modern detectors rely on closed-set supervision, where only instances from a predefined taxonomy are annotated~\cite{dhamija2020overlooked}. However, real-world datasets inevitably contain numerous unlabeled yet visually coherent object-like regions (Fig.~\ref{fig:onecol}(a)). Because these regions are treated as background during training, detectors acquire a taxonomy-dependent notion of objectness instead of learning generic object localization. Although effective for category-specific detection, this supervision paradigm fundamentally limits class-agnostic localization, whose goal is to localize objects regardless of semantic category. We refer to this limitation as {\em objectness bias}, a central challenge for class-agnostic detection under real-world supervision.

Recent approaches~\cite{kim2022learning,huang2022good,singh2025improving,wang2022open} attempt to mitigate missing annotations through heuristics such as pseudo-labeling, proposal filtering, ignore regions, and depth or grouping cues. However, these methods do not address the underlying issue: objectness is still formulated through foreground-background discrimination and therefore inherits objectness bias. Ignore-region methods~\cite{wu2018soft,niitani2019sampling,zhang2020brl} suppress uncertain regions during training while preserving negative supervision elsewhere. Likewise, partial-label and positive-unlabeled learning approaches~\cite{yang2020object} continue to rely on negative evidence over unlabeled regions.

We propose \emph{Background-Free Objectness Learning for Class-Agnostic Detection} (B-FOR), which casts class-agnostic detection as dense objectness field prediction through object-center score maps together with width and height fields across multiple scales (Fig.~\ref{fig:onecol}(b)). Object instances emerge as local spatial structures in the predicted fields, and are recovered as local maxima decoded into bounding boxes using the corresponding scale fields.

B-FOR revisits the supervision paradigm itself by learning objectness from spatially structured supervision confined to reliable annotated regions, while treating unlabeled locations as unobserved rather than explicit background negatives. This shifts the objective away from foreground-background discrimination towards learning a class-independent notion of objectness defined only where reliable supervision is available. In contrast, methods such as FCOS~\cite{tian2019fcos} and CenterNet~\cite{zhou2019objects} learn objectness through dense foreground-background supervision induced by annotated objects.

Beyond center localization, the model also learns the spatial extent associated with each candidate center, encouraging scale-consistent object representations. The combination of localized spatial supervision and scale-consistent decoding prevents degenerate solutions such as spatially diffuse objectness responses.

We evaluate our approach on unseen-category and cross-dataset class-agnostic  benchmarks~\cite{jaiswal2021class,kim2022learning}.
Our formulation leads to improvements in recall, particularly on unseen objects, with gains of up to $+10$ AR@1000 and a 2$\times$ improvement in AR@100 at IoU $\leq$ 0.5. Moreover, our approach improves cross-dataset generalization beyond annotated categories.

Our contributions are as follows:
\begin{itemize}
    \item We revisit objectness learning under incomplete annotations and identify objectness bias as a key limitation of closed-set supervision for class-agnostic detection. 
    \item We propose B-FOR, a background-free objectness learning framework that confines local spatially structured supervision to reliable annotated object regions while treating unlabeled locations as unobserved.
    \item We introduce a dense scale-field formulation that jointly models object-center evidence and spatial object extent, enabling spatially coherent object hypotheses to emerge from the learned objectness field.
\end{itemize}

\section{Related Work}
\label{sec:related}
Prior work either (i) relies on closed-set background supervision, (ii) learns objectness from sparse proposals or pseudo-labels, or (iii) mitigates missing annotations while preserving a foreground-background objective. 
In contrast, we revisit the supervision formulation itself and learn dense objectness without assigning background labels to unlabeled regions.

\textbf{Closed-set object detection.}
Modern object detectors differ substantially in architecture but share a common supervision paradigm: annotated instances define positives, while unmatched anchors, pixels, proposals, or queries are treated as background or no-object. This formulation underlies two-stage detectors~\cite{ren2016faster}, one-stage dense detectors~\cite{lin2017focal}, anchor-free methods~\cite{tian2019fcos}, center-based detectors~\cite{law2018cornernet,zhou2019objects,tian2019fcos}, and set-prediction transformers~\cite{carion2020end}.

For fully annotated closed-set benchmarks, this supervision rule is a reasonable approximation. Under incomplete annotations, however, it becomes a systematic source of objectness bias, and
detectors learn objectness conditioned on the annotated label space rather than category-independent objectness. 

\textbf{Class-agnostic objectness and proposal learning.}
Class-agnostic detection and proposal methods aim to reduce dependence on semantic labels by learning category-independent localization cues. Selective Search~\cite{uijlings2013selective}, EdgeBoxes~\cite{zitnick2014edge}, and BING~\cite{cheng2014bing}, rely on hand-crafted grouping and edge cues. DeepBox~\cite{kuo2015deepbox} and OLN~\cite{kim2022learning} estimate objectness or localization quality from candidate regions, and~\cite{jaiswal2021class} extend these ideas to full detection pipelines.

Moreover, OLN replaces the Faster R-CNN classification head with a localization-quality head based on centerness and IoU~\cite{kim2022learning}. GOOD recovers pseudo-labels for unlabeled objects using geometric cues~\cite{huang2022good}, BOWL models reliable background regions to reduce false positives from noisy pseudo-labels~\cite{singh2025improving}, and GGN learns grouping cues for category-agnostic mask proposals~\cite{wang2022open}. 
Despite their differences, these methods still rely on sparse proposals, pseudo-labels, or background-aware objectives, and do not fully address the objectness bias. 

More recently, MAVL~\cite{maaz2022class} and DiPEx~\cite{lim2024dipex} use language priors to learn class-agnostic objectness through multimodal prompts for open-vocabulary recognition. Thus, their notion of objectness remains coupled to pretrained representations.

LOST~\cite{simeoni2021localizing}, TokenCut~\cite{wang2023tokencut} and CutLER~\cite{wang2023cut} exploit self-supervised or segmentation cues to identify object regions, while PROB~\cite{zohar2023prob} models uncertainty-aware objectness for open-world detection. These methods show that objectness can emerge without category supervision, but do not address background supervision under incomplete annotations.

\textbf{Missing annotations, sparse labels, and ignore-region training.}
Missing-annotation detection methods recognize that unlabeled objects may be treated as background. Soft Sampling~\cite{wu2018soft} down-weights uncertain RoI gradients, sparse-annotation sampling~\cite{niitani2019sampling} modifies negative sampling under sparse labels, and Background Recalibration Loss~\cite{zhang2020brl} recalibrates background supervision when missing instances are optimized as negatives. Yang \etal~\cite{yang2020object} further formulate detection as a positive-unlabeled learning problem, where unlabeled regions mix true background and missing objects. More recent methods, including SparseDet~\cite{suri2023sparsedet} and S$^2$Teacher~\cite{lin2026s2teacher}, recover missing labels through pseudo-positive mining and self-training.
These methods improve robustness by reducing unreliable negatives or recovering positives. However, they largely preserve the foreground-background formulation, where unlabeled regions still contribute directly or indirectly to background modeling.

In contrast, class-agnostic localization cannot safely interpret unlabeled regions as background. B-FOR therefore removes explicit background supervision over unlabeled regions from the objectness objective and instead learns category-independent localization cues through spatially structured supervision confined to reliable annotated regions.

\begin{figure}[t]
\centering
\includegraphics[width=0.80\linewidth]{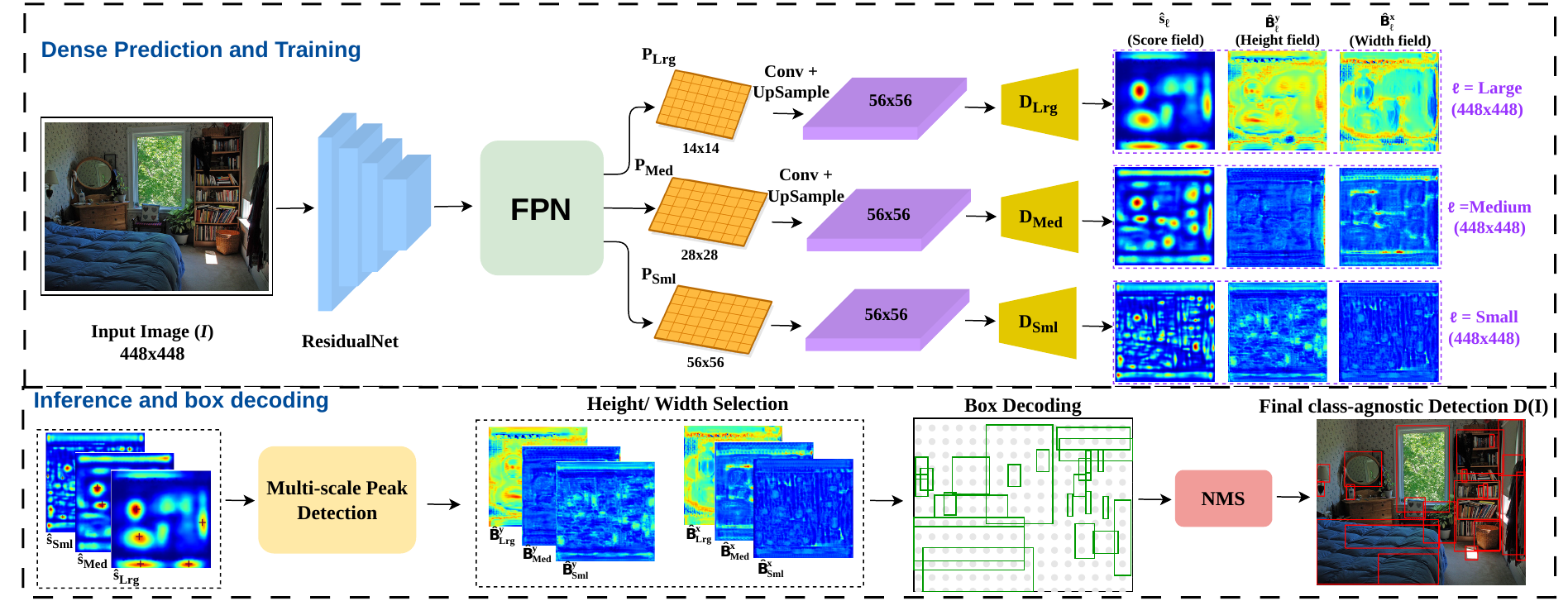}
    \caption{ Overview of B-FOR. 
Given an input image, cascaded residual and feature pyramid networks generate multi-scale feature maps that are decoded by lightweight heads $D_\ell$ into dense object representations: an object-center score field $\hat{s}_\ell$ and center-conditioned width/height fields $\hat{B}^x_\ell$ and $\hat{B}^y_\ell$ across small, medium, and large object scales. During inference, local maxima of the score fields define candidate centers, bounding boxes are recovered from the corresponding scale fields, and NMS produces the final class-agnostic detections. }
\label{fig:proposed}
\end{figure}

\section{Method}
\label{sec:ProposedMethod}
We formulate class-agnostic detection as dense prediction of multi-scale object-center evidence under incomplete supervision. 
Unlike standard dense detectors, the score field is not learned through foreground-background classification; supervision is defined only where annotated object evidence is reliable.  
The model is trained to produce spatially localized object-center responses, from which valid detections emerge as peaks in the predicted field.

Let $I \in \mathbb{R}^{H_0 \times W_0 \times 3}$ define an input image, where $H_0$ and $W_0$ denote the image height and width, respectively, and let $\Omega_0 = \{1,\ldots,W_0\} \times \{1,\ldots,H_0\}$ represent the image grid. Although the image may contain an unknown number of objects, incomplete annotation provides labels for only a subset of them. 
We predict object-center responses at three discrete object scales $\mathcal{L}=\{\text{small},\text{medium},\text{large}\}$, defined according to object size.

Let $q=(x,y)\in\Omega_0$ denote a pixel location.
For each $q \in \Omega_0$, the model predicts a scalar response $\hat{s}_{\ell}(q;I)\in[0,1]$ encoding object-center evidence at scale $\ell$. Predictions are independent across spatial locations and scales, without global normalization, allowing multiple locations to simultaneously attain high responses at the same scale.

For each location $q$ and scale $\ell$, the model predicts a center-conditioned scale field, $S_{\ell}(q;I)=
(\hat{B}^x_{\ell}(q;I),\hat{B}^y_{\ell}(q;I))\in[0,1]^2$,
encoding the normalized width and height associated with the candidate object center $q$. As a result, objectness is represented not only through localized center evidence, but also through the spatial extent occupied by the object. Each location $q=(x,y)$ at scale $\ell$ defines the candidate bounding-box
\begin{equation}
\hat{b}_{\ell}(q)= \left[x,\,y,\,W_0\hat{B}^x_{\ell}(q;I),\,H_0\hat{B}^y_{\ell}(q;I)\right].
\end{equation}

At inference time, B-FOR converts dense predictions into a ranked set of class-agnostic boxes by extracting local maxima from the object-center score fields $\hat{s}_{\ell}$ at each scale level $\ell$. 

\subsection{Architecture}
To demonstrate the proposed idea and isolate the effect of the proposed objectness learning formulation, we adopt an encoder-decoder architecture with a Feature Pyramid Network (FPN) for multi-scale prediction. The decoder uses separate heads to estimate the object-center score fields $\hat{s}_{\ell}$ and corresponding scale fields $S_{\ell}$ at each pyramid level $\ell$. An overview of the architecture is shown in Fig.~\ref{fig:proposed}. Our model counts about $31.0$M parameters.

{\bf Encoder:} We use the first five residual blocks of a ResNet-50 backbone together with an FPN to extract multi-scale feature maps $P_\ell$. Feature maps for small, medium, and large objects are derived from the outputs of the third, fourth, and fifth residual blocks, respectively.

Each feature map is defined as
$P_\ell \in\mathbb{R}^{H_\ell\times W_\ell\times C}$, with $C=128$. The FPN combines top-down and lateral connections to produce multi-scale feature representations. Before decoding, all pyramid features are resized to
a common stride-$8$ resolution ($56\times56$ for $448\times448$ inputs).
All models are trained entirely from scratch, without class-supervised pre-training, to avoid introducing semantic biases and isolate the effect of the proposed objectness supervision from category-level semantic priors.

\begin{table}[h]
\setlength{\tabcolsep}{3pt}
\begin{minipage}[t]{0.517\textwidth}
\vspace{0pt}
\textbf{Decoder:} For each pyramid level $\ell$, we attach a lightweight decoder $D_\ell$ that upsamples the feature map $P_\ell$ to the prediction grid $\Omega_0$ and predicts the fields
$\hat{s}_\ell, \hat{B}^x_\ell, \hat{B}^y_\ell$.
As shown in Fig.~\ref{fig:decoder}, the decoder alternates four $3{\times}3$ convolutional layers with ReLU activations and three $2{\times}$ upsampling stages until reaching the target resolution $H_0 \times W_0$ ($H_0{=}W_0{=}448$).
To preserve spatial detail, a stride-$8$ skip feature map is upsampled and fused with the decoder via element-wise addition before the final prediction heads. Each head consists of $3{\times}3$ convolutions followed by a $1{\times}1$ convolution and sigmoid activation.
\end{minipage}
\hfill
\begin{minipage}[t]{0.467\textwidth}
\vspace{0pt}
\centering
\includegraphics[width=0.75\linewidth]{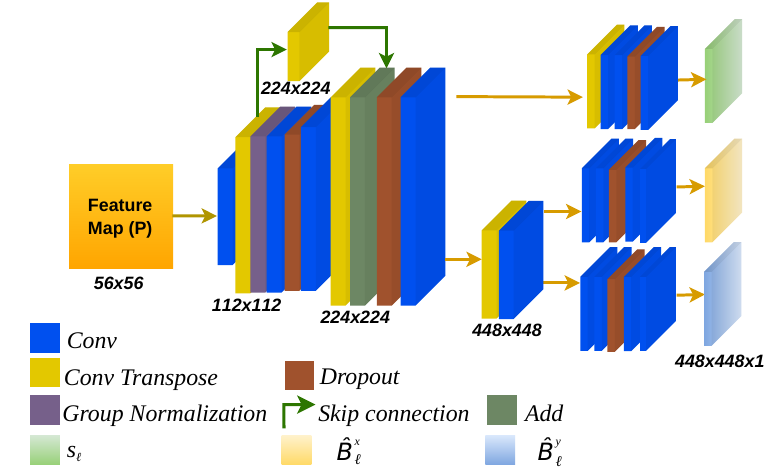}
\captionof{figure}{Decoder head for pyramid level $\ell$. Feature maps are upsampled to the prediction grid and decoded into dense outputs: an object-center score field $\hat{s}_\ell$, and normalized width and height fields, $\hat{B}^x_\ell$ and $\hat{B}^y_\ell$.}
\label{fig:decoder}
\end{minipage}
\end{table}

\subsection{Detection from Dense Object-Center Fields}
\label{sec:detection}
At inference time, B-FOR extracts a ranked set of object candidates from the predicted object-center and scale fields. Candidate centers are identified by applying local non-maximum suppression to the object-center score maps. Let $\mathcal{N}_r(q)$ denote an $r\times r$ neighborhood centered at $q$. A location $q$ is retained as a candidate center at scale $\ell$ only if its score is a local maximum within $\mathcal{N}_r(q)$ and exceeds a level-specific threshold $\tau_\ell(I)$:
\begin{equation}
\hat{s}_\ell(q;I) \geq \hat{s}_\ell(q';I)
\quad \land \quad
\hat{s}_\ell(q;I) \geq \tau_\ell(I),
\quad \forall q' \in \mathcal{N}_r(q).
\end{equation}

In our implementation, $\tau_\ell(I)$ is set to the $60$-th percentile of $\hat{s}_\ell(\cdot;I)$. This adaptive threshold acts as a pre-filter, since score fields are not globally normalized across images or pyramid levels, avoiding bias toward scales with systematically higher responses. The percentile is selected on the validation set to suppress low-score fluctuations before top-$K$ ranking. 
We retain the top-$K$ peaks per level ranked by $\hat{s}\ell(q;I)$. For each peak $q=(x,y)$ at level $\ell$, a box of minimum size $(w{\min}, h_{\min})$ is decoded from the corresponding scale-field values:
\begin{equation}
w_q=\max\big(W_0\hat{B}^x_\ell(q;I), w_{\min}\big), \qquad
h_q=\max\big(H_0\hat{B}^y_\ell(q;I), h_{\min}\big).
\end{equation}
The resulting candidate box is
$\hat{b}_\ell(q)=\left[x,\,y,\,w_q,\,h_q\right]$,
and is clipped to the image boundaries.

Candidates from all pyramid levels are merged into a single ranked list, followed by non-maximum suppression (NMS) with IoU threshold $0.5$. The final output consists of the top $D_{\max}$ detections after NMS. Unless otherwise stated, we use $r=9$,$K=1200$, $w_{\min}=h_{\min}=6$ pixels, and $D_{\max} \in \{1,10,100,1000\}$ depending on the evaluation protocol.

Importantly, local-maxima extraction, thresholding, top-$K$ selection, and NMS are applied only at inference time and are not part of training.

\section{Learning Objectness}
Training a class-agnostic detector under incomplete annotations requires exploiting reliable supervision while avoiding erroneous background penalties in unlabeled regions.

Let $\mathcal{B}=\{b_n\}_{n=1}^{N}$ denote the set of available annotations, where each box $b_n=(c_n,w_n,h_n)$ is defined by its center $c_n=(x_n,y_n)$ and spatial extent $(w_n,h_n)$. We define
$\Omega(\mathcal{B})=\bigcup_{n=1}^{N}\Omega(b_n)$
as the set of image locations covered by the annotated boxes in image $I$.

Unlike prior methods, our objectness loss is evaluated only over $\Omega(\mathcal{B})$; locations outside annotated boxes are excluded from supervision and never treated as explicit negatives. This is crucial under incomplete annotation, where the absence of a box does not imply the absence of an object.
Each box $b_n$ supervises a single decoder branch according to its scale, following the standard COCO partition based on box area.
Our training objective jointly models object-center evidence and spatial object extent:
\begin{equation}
\mathcal{L}=\mathcal{L}_{\mathrm{obj}}+\mathcal{L}_{\mathrm{SF}}.
\end{equation}

\subsection{Region-Supervised Objectness Learning}
\label{sec:ros_loss}
Conventional dense objectness losses implicitly assume exhaustive annotations by assigning low objectness scores to all unlabeled locations, thereby treating potentially unlabeled objects as negatives. For example, CenterNet~\cite{zhou2019objects} represents annotated centers with Gaussian targets and optimizes a modified Focal loss over the entire image grid, supervising all non-annotated locations as background. Similarly, FCOS~\cite{tian2019fcos} combines dense Focal loss with IoU regression while likewise treating unlabeled regions as negatives.

In contrast, the proposed Region-Supervised Objectness loss, $\mathcal{L}_{\mathrm{ROS}}$, does not define objectness through dense foreground-background supervision over the image domain. Instead, annotated boxes specify the regions where object-center supervision is reliable, while locations outside these regions are treated as unobserved with respect to the objectness objective.

This formulation differs conceptually from ignore-region training, where ambiguous regions are heuristically masked while remaining unmatched locations are still supervised as background. In contrast, $\mathcal{L}_{\mathrm{ROS}}$ reformulates the objectness objective itself by removing explicit background supervision from the loss.

{\bf Supervision signal: }
Following~\cite{zhou2019objects}, for each annotated box $b_n$, we define a Gaussian soft target $Y_n$ centered at $c_n=(x_n,y_n)$. The target is evaluated for all $q\in\Omega(b_n)$:

\begin{equation}
Y_n(q)=\exp\left(-\frac{1}{2}(q-c_n)^\top \Sigma_n^{-1}(q-c_n)\right) \qquad \text{with} \quad 
\Sigma_n = \alpha
\begin{bmatrix}
w_n^2 & 0\\
0 & h_n^2
\end{bmatrix},
\end{equation}
where $\alpha$ controls the Gaussian sharpness. The target encodes proximity to the object center, providing smooth localization supervision within annotated regions rather than object-versus-background labels.

{\bf Loss Formulation:}
Similarly to CenterNet, we optimize the objectness field using a binary cross-entropy objective with soft spatial targets. However, the targets do not encode foreground-background labels, but continuous spatial proximity to annotated object centers. The proposed Region-Supervised Objectness loss is defined as
\begin{equation*}
\begin{split}
\mathcal{L}_{\text{ROS}}=\frac{1}{N}\sum_{n=1}^{N}
\frac{1}{|\Omega(b_n)|}\sum_{(x,y)\in\Omega(b_n)}\Big[
- Y_n(x,y)\log \hat{s}_{\ell_n}(x,y)
-(1-Y_n(x,y))\log\big(1-\hat{s}_{\ell_n}(x,y)\big)\Big],
\end{split}
\end{equation*}
where $\ell_n$ denotes the scale associated with box $b_n$. Normalization by $|\Omega(b_n)|$ ensures that each object contributes equally to the loss, independently of its spatial extent.

Although the objectness objective includes the complementary term $(1-Y_n)$, supervision remains confined to annotated object regions and therefore does not impose background constraints on unmatched image locations. Instead, it induces local spatial competition within reliable support, encouraging peaked responses around object centers and progressively lower responses in surrounding regions. As a result, the model learns a spatially structured objectness representation without explicit negative supervision outside annotated regions.

However, our regional objective may produce slightly shifted local maxima due to its independent optimization at each spatial location, particularly for large or partially overlapping objects. To stabilize center localization, we introduce a center-consistency regularizer that aligns the first moment of the predicted regional score distribution with the annotated object center. This term does not introduce additional objectness supervision; instead, it regularizes the spatial structure of the predicted score field from which object hypotheses are decoded.

Specifically, we estimate the predicted object center $\hat{c}_n=(\hat{x}_n,\hat{y}_n)$ by applying a differentiable soft-argmax operator to the predicted score map $\hat{s}_{\ell_n}$ restricted to $\Omega(b_n)$ in prediction-grid coordinates $\Omega_0$. Normalization by $|\Omega(b_n)|$ balances the contribution of objects across scales. The resulting center-consistency regularization term is
\begin{equation}
\mathcal{L}_{\text{ctr}} = \frac{1}{N}\sum_{n=1}^{N}
\frac{1}{|\Omega(b_n)|}
\|\hat{c}_n - c_n \|^2_2.
\end{equation}

By introducing $\lambda_{\mathrm{ctr}}$ to control the strength of the center-consistency regularization, we define the final objectness objective $\mathcal{L}_{\mathrm{obj}}$ as
\begin{equation}
\mathcal{L}_{\mathrm{obj}} =
\mathcal{L}_{\mathrm{ROS}}
+\lambda_{\mathrm{ctr}}\mathcal{L}_{\mathrm{ctr}}.
\end{equation}

\subsection{Learning Dense Scale Fields}
\label{sec:scale_loss}
Standard center-based detectors regress object size only at annotated object centers. In contrast, B-FOR allows detections to emerge from any local maximum of the learned objectness field. The model must therefore learn not only object-center evidence, but also the bounding-box geometry associated with each candidate location. This reflects a fundamental property of objectness: objects are spatially extended structures rather than isolated center responses.

To address this issue, we learn displacement-aware scale fields. For each candidate location $q$, the model predicts the width and height of the box that would enclose the object if decoded from $q$. Consequently, supervision is not tied to a single target at the annotated object center; instead, object extent becomes a spatially varying quantity determined by both candidate location and object scale.
\begin{table}[h!]
\setlength{\tabcolsep}{3pt}
\begin{minipage}[h!]{0.49\textwidth}
\vspace{0pt}
Specifically, dense scale fields are estimated independently at each scale $\ell$, producing prediction maps $\hat{B}^x_\ell,\hat{B}^y_\ell \in [0,1]^{H_0\times W_0}$ encoding the normalized width and height of candidate boxes decoded from each pixel location.

Consider a candidate center $q=(u,v)$. As shown in Fig.~\ref{fig:box_loss}, if the annotated object center is $c_n=(x_n,y_n)$, a box centered at $q$ can still enclose the object by expanding to compensate for the center displacement. Let $w_n$ and $h_n$ denote the annotated object width and height, the minimum width and height satisfying this condition are
\begin{eqnarray}
w_{q} &=& w_n + 2|x_n-u|,\\
h_{q} &=& h_n + 2|y_n-v|.
\end{eqnarray}

\end{minipage}
\hfill
\begin{minipage}[h!]{0.49\textwidth}
\centering
\includegraphics[width=0.8\linewidth]{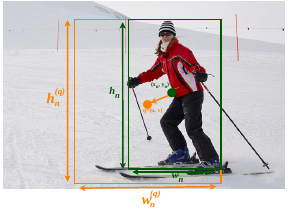}
   \captionof{figure}{The green box denotes the annotated object box $b_n=(x_n,y_n,w_n,h_n)$. When a box is decoded at $q=(u,v)$ and must enclose the object, its extent expands symmetrically to compensate for the offset. The orange box, centered at $q$, has width $w_q$ and height $h_q$.}
   \label{fig:box_loss}

\end{minipage}
\end{table}

For a pixel location $q$, the scale-field regression loss is
\begin{equation}
\mathcal{L}(q|b_n) = \left|\hat{B}^x_{\ell_n}(q;I) -\frac{w_{q}}{W_0} \right| + \left|
\hat{B}^y_{\ell_n}(q;I) - \frac{h_{q}}{H_0}\right|,
\end{equation}
where normalization by image size matches the target range $[0,1]$.

We do not supervise the scale field over the entire annotated box, as this would overemphasize highly off-center locations and increase interference between nearby or overlapping objects. Instead, supervision is restricted to compact $k\times k$ neighborhoods $\mathcal{N}_k(\cdot)$ centered at the annotated object center $c_n=(x_n,y_n)$ and the predicted center $\hat{c}_n=(\hat{x}_n,\hat{y}_n)$ obtained via differentiable soft-argmax over the object-center score map within $\Omega(b_n)$.
The neighborhood around $c_n$ provides a stable geometric anchor for learning object extent, while the neighborhood around $\hat{c}_n$ reduces train-test mismatch by supervising the locations from which object hypotheses are decoded at inference time.

The scale-field loss for box $b_n$ is
\begin{equation}
\begin{aligned}
\mathcal{L}_{b_n} =
\frac{1}{k^2}
\sum_{q\in\mathcal{N}_k(\hat{x}_n,\hat{y}_n)}
\mathcal{L}(q|b_n)+\frac{1}{k^2}
\sum_{q\in\mathcal{N}_k(x_n,y_n)}
\mathcal{L}(q|b_n).
\end{aligned}
\label{sf_equation}
\end{equation}
We note that overlaps between supervision neighborhoods are rare, as distinct objects are expected to induce separate local maxima in the object-center field. Significant overlap would indicate ambiguous or insufficiently separated peaks in the learned objectness map.

The image-level scale-field loss is
\begin{equation}
\mathcal{L}_{\mathrm{SF}}
=
\frac{1}{N}\sum_{n=1}^{N}\mathcal{L}_{b_n}.
\end{equation}

Together with the proposed object-center supervision, the dense scale fields allow object hypotheses to emerge as spatially coherent structures rather than isolated center activations. Unlike CenterNet, which regresses object size only at annotated centers, or FCOS, which relies on foreground-assigned regression targets, our formulation learns object geometry and extent directly from candidate centers emerging in the learned objectness field.

\section{Experimental Setup and Results}
\label{sec:experiments}

\textbf{Datasets.} 
We use three standard detection benchmarks: 
(1) PASCAL VOC 2007+ 2012~\cite{everingham2010pascal}, containing $\sim$27k images annotated with 20 categories; 
(2) MS-COCO 2017~\cite{lin2014microsoft}, with $\sim$120k images across 80 categories;  
(3) Open Images~\cite{kuznetsova2020open}, comprising over 570 categories.
OpenImages is used only to evaluate generalization to unseen objects. 

Under fully class-agnostic object detection protocols, class identities are never used. We consider two settings:

\textbf{(i) Generalization to unseen categories}~\cite{jaiswal2021class}, where evaluation is performed exclusively on unseen object categories while ignoring seen categories. This setting includes:
\begin{itemize}
 \item \textbf{Seen/Unseen split:} training on 17 VOC classes and evaluating on 3 unseen classes.

\item\textbf{Cross-dataset unseen-category evaluation:} training on one dataset and evaluating on disjoint categories from another. We train on VOC20 and evaluate on COCO60, and train on COCO80 while evaluating on 490 Open Images categories.
\end{itemize}

\textbf{(ii) Cross-dataset generalization,} where the model is trained on one dataset and evaluated on another containing both seen and unseen categories. Specifically, we train on VOC and evaluate on COCO, and train on COCO while evaluating on VOC and Open Images.

\textbf{Evaluation Metrics.}
Since the proposed model is class-agnostic, it may correctly detect valid objects that are not annotated in the evaluation set. In this setting, standard mean Average Precision (mAP) can be misleading, as such detections would be counted as false positives. Following the protocol in~\cite{jaiswal2021class}, we report average recall AR@IoU$=0.5$, which directly measures localization performance independently of semantic category. Recall is reported overall and across object-size groups (small, medium, and large).

\textbf{Implementation details.}
We implement the proposed method in TensorFlow without external detection frameworks. Input images are zero-padded to a square canvas, resized to $448 \times 448$, and normalized to $[0,1]$. Annotated boxes are mapped to resized image coordinates and assigned to scale levels according to the square root of their area, following the standard COCO partition.
To control memory usage and maintain balanced training batches, we randomly subsample at most 10 annotated instances per image during training. We use $k=9$ in Eq.~\ref{sf_equation} for scale-field supervision.

Optimization uses Adam with an initial learning rate of $10^{-4}$. The learning rate is reduced by a factor of $0.3$ after 3 epochs without validation improvement, down to a minimum of $10^{-9}$. Training employs early stopping with patience 10 based on validation loss. All experiments are conducted on a single NVIDIA GeForce RTX 4090 GPU.

\subsection{Results}
\textbf{(i) Generalization to unseen categories.}
We first evaluate B-FOR on the VOC 17/3 seen-unseen split. 
Results in Table~\ref{tab:voc_seen_unseen} are reported for both the single-scale variant and the full multi-scale FPN-based model. 
B-FOR with FPN achieves $AR_{Ovr}=80.3$, outperforming the Faster R-CNN based class-agnostic baseline ($60.1$). 
Although the SSD-based baseline achieves higher recall on the  \emph{cow} and \emph{boat} categories, B-FOR obtains superior performance on the \emph{Tvmonitor} category, achieving the highest recall of $92.2$ and higher overall recall. 
The FPN-based model improves over the single-scale variant across all metrics, highlighting the importance of multi-scale object-center representations for unseen-category localization.

Across individual categories (Table~\ref{tab:per_class_recall}), recall is particularly high for \emph{Tvmonitor}, while \emph{boat} remains more challenging at small and medium scales.

\begin{table}[t!]
\begin{center}
\footnotesize
\setlength{\tabcolsep}{3pt}
\begin{minipage}{0.48\textwidth}
\begin{center}
\begin{tabular}{|l|c|c|c|c|}
\hline
\textbf{Model} & \textbf{Ovr} & \textbf{Cow} & \textbf{Boat} & \textbf{Tvmonitor} \\
\hline\hline
SSD-ag-ad~\cite{jaiswal2021class} & 72.2 & \textbf{83.3} & \textbf{68.9} & 66.0 \\
FRCNN-ag-ad~\cite{jaiswal2021class} & 60.1 & 82.3 & 53.7 & 47.9 \\
B-FOR (Ours) (w/o FPN) & 37.3 & 46.2 & 29.2 & 37.9 \\
B-FOR (Ours) & \textbf{80.3} & 75.3 & 66.6 & \textbf{92.2}  \\
\hline
\end{tabular}
\end{center}
\caption{\textbf{Seen/unseen split.} Class-agnostic training on 17 PASCAL VOC classes and evaluation on the remaining 3 unseen classes using AR@1000 at IoU $\geq 0.5$.}
\label{tab:voc_seen_unseen}
\end{minipage}
\hfill
\begin{minipage}{0.48\textwidth}
\begin{center}
\begin{tabular}{|l|c|c|c|}
\hline
\textbf{Class} & \textbf{AR$_{Sml}$} & \textbf{AR$_{Med}$} & \textbf{AR$_{Lrg}$} \\
\hline\hline
Cow (Easy) & 32.5 & 78.4 & 85.0 \\
Boat (Medium) & 31.5 & 65.4 & 89.1 \\
Tvmonitor (Hard) & 50.0 & 93.4 & 96.2 \\
\hline
Overall & 36.2 & 81.7 & 93.4\\
\hline
\end{tabular}
\end{center}
\caption{Per-class size-wise recall on the VOC seen/unseen split. AR$_{Sml}$, AR$_{Med}$, and AR$_{Lrg}$ correspond to small, medium, and large objects, respectively.}
\label{tab:per_class_recall}
\end{minipage}
\end{center}
\end{table}

\begin{table}[t]
\begin{center}
\footnotesize
\begin{tabular}{|l|c|c|c|c|c|c|c|c|}
\hline
\multicolumn{1}{|l|}{} & \multicolumn{4}{c|}{\textbf{VOC 20 $\rightarrow$ COCO 60}} & \multicolumn{4}{c|}{\textbf{COCO 80 $\rightarrow$ OI 490}} \\
\cline{2-9}
\textbf{Model} & \textbf{$AR_{Ovr}$} & \textbf{$AR_{Sml}$} & \textbf{$AR_{Med}$} & \textbf{$AR_{Lrg}$} & \textbf{$AR_{Ovr}$} & \textbf{$AR_{Sml}$} & \textbf{$AR_{Med}$} & \textbf{$AR_{Lrg}$} \\
\hline
CenterNet$^\dagger$~\cite{zhou2019objects} & 17.8 & 6.6 & 17.8 & 41.4 & 37.4 & \textbf{9.0} & 25.7 & 48.4 \\
FCOS$^\dagger$~\cite{tian2019fcos} & NA & NA & NA & NA & 41.3 & 7.5 & 24.6 & 55.4  \\
FRCNN-aw~\cite{jaiswal2021class} & 12.4 & 1.8 & 11.0 & 37.0 & 17.0 & 1.6 & 8.3 & 33.0 \\
FRCNN-ag~\cite{jaiswal2021class} & 13.6 & 2.4 & 13.1 & 38.0 & 18.2 & 2.8 & 12.5 & 31.8 \\
SSD-ag~\cite{jaiswal2021class} & 18.1 & 3.1 & 23.9 & 40.4 & 20.6 & 2.3 & 20.6 & 31.4 \\
SSD-ag-ad~\cite{jaiswal2021class} & 19.6 & 3.6 & 25.2 & 44.3 & 21.0 & 2.4 & 21.1 & 31.9 \\
FRCNN-ag-ad~\cite{jaiswal2021class} & 15.2 & 2.7 & 15.8 & 40.6 & 19.2 & 2.8 & 13.1 & 33.7 \\ 
UniDetector~\cite{wang2024unidetector} &38.1 & \textbf{16.4} & 49.6 & 65.2& NA & NA & NA & NA \\
B-FOR (Ours) (w/o FPN) & 13.3 & 0.2 & 4.4 & 55.1 & 30.9 & 0.7 & 13.4 & 44.4 \\
B-FOR (Ours) & \textbf{48.7} & 11.0 & \textbf{71.1} & \textbf{91.8} & \textbf{60.2} & 7.2 & \textbf{45.6} & \textbf{77.5} \\
\hline
\end{tabular}
\end{center}
\caption{\textbf{Cross-dataset unseen-category evaluation.} AR@1000 at IoU $\geq 0.5$; all methods evaluated class-agnostically. FRCNN, SSD, and UniDetector results are from~\cite{jaiswal2021class, wang2024unidetector};
$\dagger$~own evaluation on public checkpoints (CenterNet DLA-34, FCOS ResNet-50), included as class-supervised references, not matched
baselines. FCOS VOC20$\to$COCO60 unavailable: no public VOC-trained checkpoint.
B-FOR trains from scratch without class-level supervision.}

\label{tab:cross_dataset_unseen}
\end{table}

On the more challenging cross-dataset unseen-category benchmarks, B-FOR achieves strong gains. As shown in Table~\ref{tab:cross_dataset_unseen}, training on VOC20 and evaluating on the 60 non-overlapping COCO categories yields $AR_{Ovr}=48.7$, outperforming UniDetector~\cite{wang2024unidetector} ($38.1$) by $+10.6$ points. Gains are especially pronounced for medium and large objects, reaching $71.1$ and $91.8$ recall, respectively. For UniDetector, we report the results from their Table~XIV, where the
authors adopt the same protocol specifically to enable
comparison with~\cite{jaiswal2021class}. We also report B-FOR results without using FPN, showing it cannot
adapt its receptive field to the object scale distribution of the target domain. Without FPN, the model is biased toward the largest objects present in the training dataset.

Similarly, training on COCO80 and evaluating on 490 non-overlapping Open Images categories yields $AR_{Ovr}=60.2$, substantially improving over the previous best results.

We also compare with CenterNet and FCOS models, which are trained using class supervision but with the closed-set assumption that suppresses the unlabeled objects during training. B-FOR achieves significantly higher performance. These results are consistent with the hypothesis that objectness learned without global foreground-background supervision generalizes more effectively beyond the training taxonomy. Importantly, the improvement extends beyond held-out categories within the same dataset and persists across substantial shifts in dataset bias, taxonomy, and annotation policy. 

\begin{figure*}[t!]
  \centering   \includegraphics[width=0.70\textwidth]{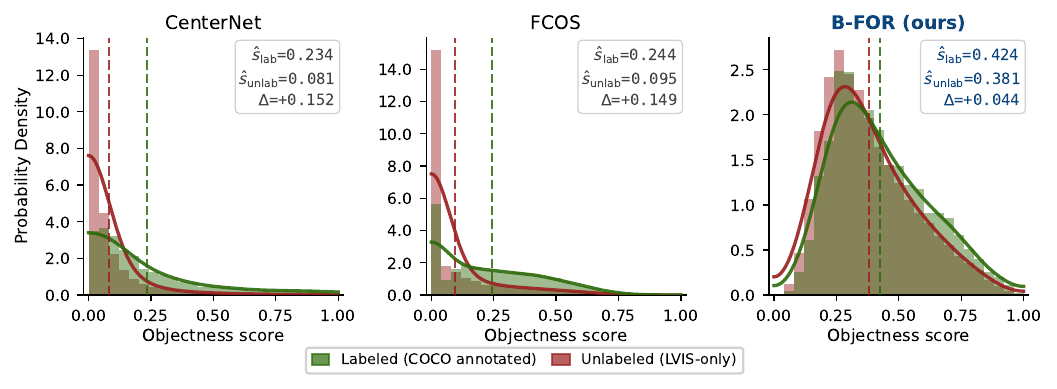}
    \caption{Objectness score distributions at annotated (\emph{labeled}, green) and unannotated (\emph{unlabeled}, red) object centers. $y$-axis: probability density, bin width~$= 0.04$.
    Dashed lines mark group means; $\Delta = \hat{s}_\text{lab} -
    \hat{s}_\text{unlab}$ quantifies objectness bias.
    CenterNet and FCOS both exhibit a sharp spike near zero for unlabeled
    objects reflecting suppression
    of unlabeled regions through foreground--background supervision. In contrast, B-FOR yields a broader unlabeled distribution with substantial overlap between labeled and unlabeled objects.}
   \label{fig:distribution}
\end{figure*}

To quantify objectness bias, we evaluate models trained on COCO80 by measuring the mean objectness score at the centers of COCO-annotated (\emph{labeled}, $n{=}4{,}999$) and LVIS-only (\emph{unlabeled}, $n{=}30{,}256$) objects absent from COCO. Fig.~\ref{fig:distribution} shows that CenterNet~\cite{zhou2019objects} and FCOS~\cite{tian2019fcos} assign substantially lower scores to unlabeled objects, producing large labeled--unlabeled score gaps ($\Delta{=}{+}0.152$ and ${+}0.149$). This reflects the effect of foreground--background supervision, where unannotated regions are treated as negatives. In contrast, B-FOR reduces the gap to $\Delta{=}{+}0.044$ ($3.4\times$ lower) and assigns a mean score of $0.381$ to unlabeled objects, which is $4.7\times$ and $4.0\times$ higher than CenterNet and FCOS, respectively. Moreover, the unlabeled score distribution of B-FOR largely overlaps the labeled one, indicating that removing explicit background supervision reduces objectness bias and allows the model to respond to valid objects regardless of annotation status. We further verify this against pseudo-background regions and provide qualitative examples of objectness bias in the supplementary material (Sec.~A, E).

\begin{figure*}[t!]
  \centering
   \includegraphics[width=0.80\textwidth]{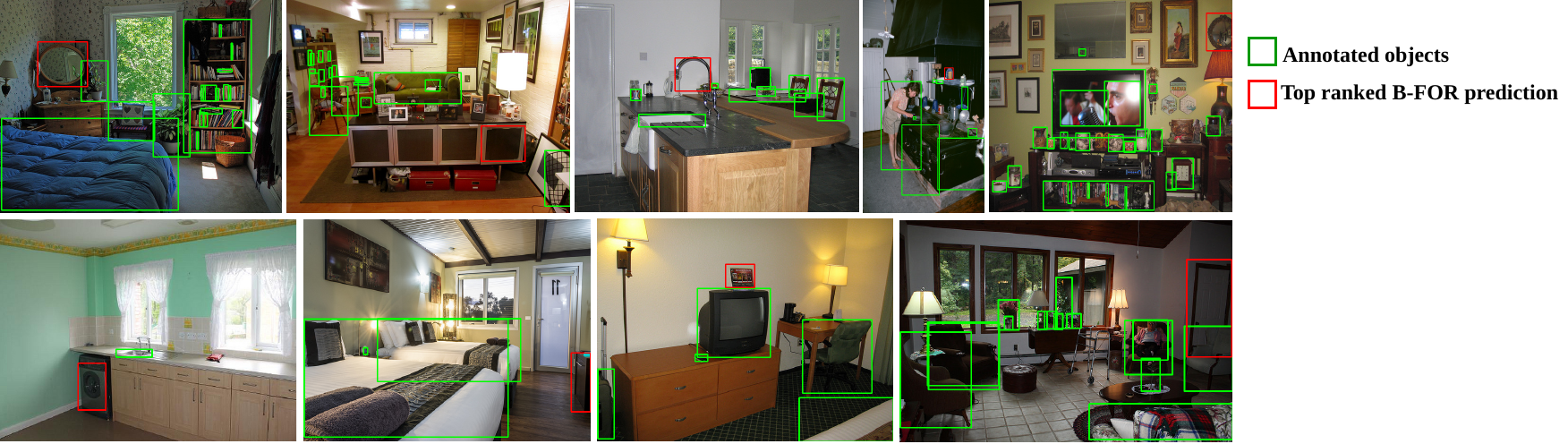}
    \caption{Top-1 CAOD on COCO \texttt{val} (VOC20-trained model). Many top-1 predictions localise visually coherent but unannotated instances outside the COCO60 evaluation subset, counted as false positives under AR@1 despite reflecting valid objectness.}
   \label{fig:top1}
\end{figure*}

\begin{table}[b!]
\centering
\footnotesize
\setlength{\tabcolsep}{4pt}
\begin{minipage}[t]{0.48\textwidth}
\centering
\begin{tabular}{|l|c|c|c|}
\hline
\textbf{Model} & \textbf{AR$_1$} & \textbf{AR$_{10}$} & \textbf{AR$_{100}$} \\
\hline\hline
FRCNN-ag~\cite{jaiswal2021class} & \textbf{7.0} & 11.0 & 13.0 \\
FRCNN-ag-ad~\cite{jaiswal2021class} & 6.5 & 10.0 & 17.0 \\
B-FOR (Ours) & 2.1 & \textbf{11.8} & \textbf{38.0} \\
\hline
\end{tabular}
\caption{Cross-dataset evaluation for VOC20$\rightarrow$COCO60. We report recall for the top-1, top-10, and top-100 detections.}
\label{tab:coco_top_dets}
\end{minipage}
\hfill
\begin{minipage}[t]{0.48\textwidth}
\centering
\begin{tabular}{|l|c|c|c|c|}
\hline
\textbf{Train $\rightarrow$ Val} & \textbf{AR$_{Ovr}$} & \textbf{AR$_{Sml}$} & \textbf{AR$_{Med}$} & \textbf{AR$_{Lrg}$} \\
\hline\hline
VOC $\rightarrow$ COCO & 53.9 & 13.8 & 75.8 & 93.5 \\
COCO $\rightarrow$ VOC & 88.3 & 58.6 & 83.0 & 97.7 \\
COCO $\rightarrow$ OI & 60.9 & 7.7 & 45.7 & 77.3 \\
\hline
\end{tabular}
\caption{Cross-dataset class-agnostic generalization. AR@1000 at IoU~$\geq 0.5$ is reported overall and by object size.}
\label{tab:cross_dataset}
\end{minipage}
\end{table}
However, B-FOR exhibits relatively low AR@1 in Table~\ref{tab:coco_top_dets}. Interpreting AR@1 under incomplete annotations is inherently difficult, since the model is trained to capture class-agnostic objectness rather than dataset-specific labels. Consequently, the top-ranked detection may correspond to a visually valid but unlabeled object instance, which standard evaluation incorrectly counts as a false positive. As illustrated in Fig.~\ref{fig:top1}, many top-1 detections (shown in red) correspond to coherent object instances missing from COCO annotations.

To quantify this effect, we manually inspect top-1 predictions on 100 random COCO \texttt{val} images. With the VOC20-trained model, 48\% correspond to valid but unlabeled objects, 32\% valid labeled objects, and 20\% are true false positives; with COCO80-trained model, the split is 42\%, 34\%, and 24\%, respectively.
Table~\ref{tab:coco_top_dets} further shows that recall increases substantially as the proposal budget grows, reaching $38.0$ AR@100. 

\textbf{(ii) Cross-dataset generalization.}
Cross-dataset evaluation tests whether the learned objectness transfers beyond the source dataset taxonomy under shifts in object categories, image distribution, annotation policy, and dataset bias.

Table~\ref{tab:cross_dataset} shows that B-FOR maintains high class-agnostic recall across these shifts. Training on VOC and evaluating on COCO yields $AR_{\mathrm{Ovr}} =53.9$, with particularly high recall on medium and large objects ($75.8$ and $93.5$, respectively). Training on COCO and evaluating on VOC further increases recall to $AR_{\mathrm{Ovr}}=88.3$. Unlike the unseen-category protocols, COCO to VOC does not introduce semantic novelty, since VOC categories are contained in COCO. Instead, it evaluates whether the learned objectness transfers across substantial shifts in dataset statistics, image complexity, and annotation policy, confirming that region-supervised objectness learning captures object
structure beyond the annotated categories. Qualitative results of B-FOR are discussed in the supplementary material (Sec.~B).

\subsection{Ablation studies}
\label{sec:ablations}
We ablate our approach on the VOC20$\rightarrow$COCO60 split while keeping the same architecture and training setup; only the objectness and scale-field losses differ. 
The ablation tests two hypotheses: (i) objectness should be learned without closed-set background supervision; (ii) effective objectness learning requires both object centers and spatial extent notions.

Table~\ref{tab:ablation_losses1} shows that the scale-field loss is critical even with 
conventional objectness supervision. $\mathcal{L}_{\mathrm{CE}}$ denotes conventional dense objectness supervision with foreground-background labels. Replacing direct $\mathcal{L}_{\mathrm{L1}}$ width/height
regression with the proposed $\mathcal{L}_{\mathrm{SF}}$ improves
$\mathrm{AR}_{\mathrm{Ovr}}$ from $20.8$ to $27.2$, with large gains for medium and large
objects. This suggests that $\mathcal{L}_{\mathrm{SF}}$
provides structured spatial extent supervision for dense center-based decoding.

\begin{table}[t!]
\footnotesize
\begin{center}
\begin{tabular}{|l|c|c|c|c|}
\hline
\textbf{Setting} & \textbf{AR$_{Ovr}$} & \textbf{AR$_{Sml}$} & \textbf{AR$_{Med}$} & \textbf{AR$_{Lrg}$} \\
\hline\hline
$\mathcal{L}_{\mathrm{CE}}$ + $\mathcal{L}_{\mathrm{L1}}$ & 20.8 & 5.2 & 29.6 & 39.7 \\
$\mathcal{L}_{\mathrm{CE}}$ + $\mathcal{L}_{\mathrm{SF}}$ & 27.2 & 5.0 & 40.8 & 52.0 \\
$\mathcal{L}_{\mathrm{obj}}$ + $\mathcal{L}_{\mathrm{L1}}$ & 18.5 & 0.7 & 22.3 & 50.0 \\
$\mathcal{L}_{\mathrm{obj}}$ + $\mathcal{L}_{\mathrm{L1}}$($\mathcal{N}_k$) & 33.4 & \textbf{12.9} & 48.1 & 46.8 \\
$\mathcal{L}_{\mathrm{obj}}$ + $\mathcal{L}_{\mathrm{SF}}$ & \textbf{48.7} & 11.0 & \textbf{71.1} & \textbf{91.9} \\
$\mathcal{L}_{\mathrm{ROS}}$ + $\mathcal{L}_{\mathrm{SF}}$ & 23.6 & 1.5 & 32.2 & 56.3 \\
\hline
\end{tabular}
\end{center}
\caption{
Ablation of supervision design on the VOC$\rightarrow$COCO60 split. $\mathcal{L}_{\mathrm{CE}}$ and $\mathcal{L}_{\mathrm{obj}}$ denote conventional (foreground vs background) and proposed objectness supervision, respectively. $\mathcal{L}_{\mathrm{L1}}$, $\mathcal{L}_{\mathrm{L1}}(\mathcal{N}_k)$, and $\mathcal{L}_{\mathrm{SF}}$ denote global regression, local-neighborhood regression, and the proposed scale-field loss. $\mathcal{L}_{\mathrm{ROS}}$ indicates $\mathcal{L}_{\mathrm{obj}}$ without $\mathcal{L}_{\mathrm{ctr}}$ term.
}
\label{tab:ablation_losses1}
\end{table}
The proposed objectness loss changes the supervision assumption rather than heuristically suppressing uncertain regions. With $\mathcal{L}_{\mathrm{obj}}$, unannotated locations are not treated as background; objectness is learned only from reliable annotated object regions. However, pairing this formulation with standard $\mathcal{L}_{\mathrm{L1}}$ regression does not improve overall recall, since the localization branch assumes boxes are decoded only from annotated centers. This mismatch is evident in
$\mathcal{L}_{\mathrm{obj}}+\mathcal{L}_{\mathrm{L1}}$, which improves large-object recall but reduces small- and medium-object recall with respect to $\mathcal{L}_{\mathrm{CE}}$.

The benefit emerges when the background-free objectness objective is paired with a compatible localization loss. Restricting $\mathcal{L}_{\mathrm{L1}}$ to local neighborhoods improves $\mathrm{AR}_{\mathrm{Ovr}}$ from $18.5$ to $33.4$, highlighting the importance of local supervision around candidate centers. The full model, $\mathcal{L}_{\mathrm{obj}}+\mathcal{L}_{\mathrm{SF}}$, achieves best performance with $\mathrm{AR}_{\mathrm{Ovr}}=48.7$. Compared with $\mathcal{L}_{\mathrm{CE}}+\mathcal{L}_{\mathrm{SF}}$, the gain is $+21.5$ points, showing the impact of removing closed-set background supervision.
Compared with $\mathcal{L}_{\mathrm{obj}}+\mathcal{L}_{\mathrm{L1}}$ neighborhood
supervision, it improves by $+15.3$ points, confirming that the displacement-aware
scale-field objective is substantially stronger than local width/height regression. These results support B-FOR's design principle: background-free objectness learning must be coupled with a center-conditioned scale field that models object extent as a function of candidate center location. $\mathcal{L}_{\mathrm{ROS}}+\mathcal{L}_{\mathrm{SF}}$ shows that removing $\mathcal{L}_{\mathrm{ctr}}$ drops $AR_{Ovr}$ to $23.6$, confirming its role in stabilizing center localization.

\subsection{Robustness of the Detection Stage}
\label{sec:limitation_inference}
B-FOR predicts dense object-center and scale fields, whereas evaluation requires a finite ranked set of boxes. Therefore, results depend not only on the learned dense representation, but also on the decoding procedure. In our implementation, local-maximum filtering, percentile thresholding, top-$K$ selection, and NMS are applied sequentially to obtain a compact detection set. While effective, these steps can suppress valid instances in crowded scenes, merge nearby objects, or discard low-scoring small objects before evaluation.

As already explained, the decoding procedure relies on the adaptive percentile threshold $\tau_\ell(I)$ to suppress low-score fluctuations before top-$K$ candidate ranking. It is therefore important to assess the impact of this threshold on overall performance.

\begin{figure}[t!]
  \centering
\includegraphics[width=0.75\textwidth]{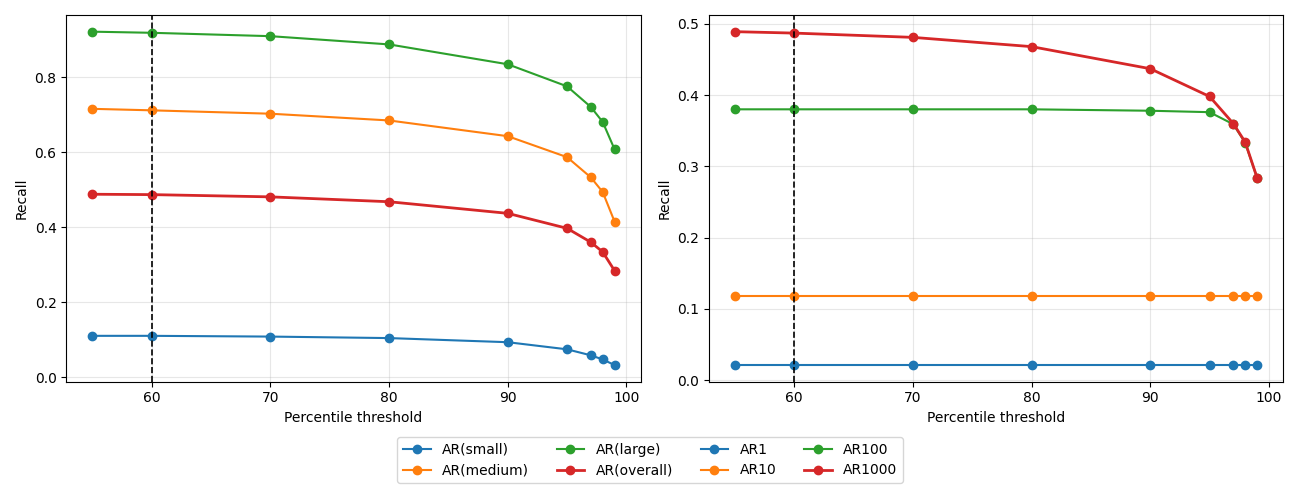}
   \caption{Effect of the percentile threshold $\tau_\ell$ on AR@1000 (IoU~$\geq 0.5$, VOC20$\to$COCO60 split). \textbf{Left:} size-wise recall (small, medium, large, overall)@1000. \textbf{Right:} recall at proposal budgets (AR@${1, 10, 100, 1000}$). The dashed vertical line marks the chosen 60th percentile. Recall remains stable up to the 80th percentile and degrades sharply beyond 90th.}
   \label{fig:percentile}
\end{figure}

All results below are reported on the VOC20$\to$COCO60 validation split.
Fig.~\ref{fig:percentile} shows that B-FOR is remarkably robust to the choice of the percentile threshold $\tau_\ell$. Across both size-wise recall metrics and proposal budgets (AR@1-1000), performance remains essentially unchanged over a wide operating range, up to approximately the 80th percentile. Only extremely aggressive thresholding (beyond the 90th percentile) leads to a noticeable degradation, as informative object responses begin to be suppressed.
These results indicate that the adaptive threshold acts primarily as a denoising mechanism rather than a critical hyperparameter, making the decoding procedure stable and easy to tune in practice. Based on this analysis, we use the 60th percentile threshold in all experiments.

\section{Conclusion}
We presented B-FOR, a dense class-agnostic detection framework that lessens objectness bias: the systematic suppression of unlabeled objects caused by foreground-background supervision in closed-set detectors. Rather than compensating for this bias through heuristics, B-FOR avoids explicit background supervision outside ground-truth boxes altogether, learning objectness exclusively from reliable annotated regions using spatially structured soft targets. Object hypotheses emerge as local maxima of predicted multi-scale object-center fields, decoded into bounding boxes through displacement-aware scale fields that model object extent as a spatially varying function of candidate center location.

Experiments on PASCAL VOC, MS-COCO, and Open Images demonstrate strong generalization to unseen categories and across dataset shifts. B-FOR achieves $AR_{Ovr}$ = 80.3 on the VOC seen/unseen split, +10.6 AR@1000 over UniDetector on VOC20$\rightarrow$COCO60, and $AR_{Ovr}$ = 60.2 on COCO80$\to$OpenImages490, all without ImageNet pretraining. Objectness bias is reduced $3.4\times$ relative to CenterNet and FCOS ($\Delta{=}{+}0.044$ vs. ${+}$0.152 and ${+}$0.149). Ablation studies show that the full $\mathcal{L}_{\mathrm{obj}}$+$\mathcal{L}_{\mathrm{SF}}$ formulation improves $AR_{Ovr}$ by ${+}$21.5 points over conventional supervision with scale fields ($\mathcal{L}_{\mathrm{CE}}+\mathcal{L}_{\mathrm{SF}}$), and by ${+}$15.3 points over background-free supervision with local $\mathcal{L}_{\mathrm{L1}}$ regression, confirming that both components are essential. These results provide evidence that objectness is fundamentally a spatially distributed property, arising from extended object regions rather than single center responses.

A current limitation of our work is the decoding stage: local-maximum extraction and NMS are not jointly optimized with the learned fields, limiting low-budget recall such as AR@1. Learning structured decoding directly from dense objectness fields is a key direction for future work. Another promising direction is extending the proposed supervision formulation beyond bounding-box detection toward class-agnostic instance segmentation. Since B-FOR already represents objectness as a spatially structured field, the formulation may naturally generalize to dense mask prediction and open-world segmentation.

\clearpage
\appendix
\section*{Supplementary Material}
\addcontentsline{toc}{section}{Supplementary Material}

\section{Objectness Bias Visualization}
\label{sec:objectness_bias}
\begin{figure}[t!]
 \centering
\includegraphics[width=\textwidth]{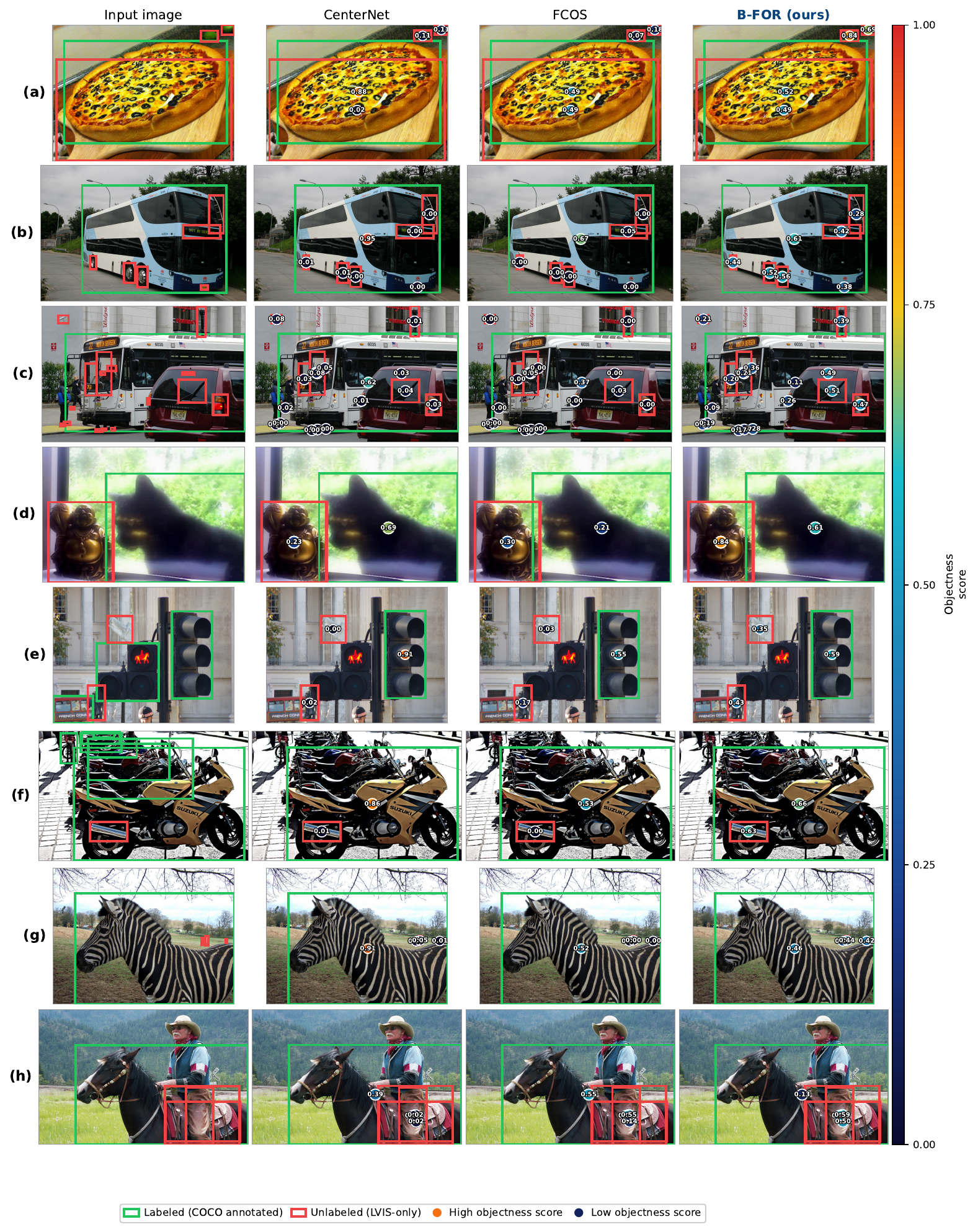}
   \caption{Objectness bias visualization on COCO \texttt{val2017}. Green boxes denote labeled COCO objects, while red boxes indicate objects unlabeled during training. Filled circles show objectness scores, color-coded from low (\textcolor[HTML]{1a3a8a}{$\bullet$}) to high (\textcolor[HTML]{d62728}{$\bullet$}). CenterNet~\cite{zhou2019objects} and FCOS~\cite{tian2019fcos} assign consistently low scores to unlabeled objects, revealing objectness bias caused by foreground-background supervision. In contrast, B-FOR assigns similar scores to both labeled and unlabeled objects, substantially reducing objectness bias.
   }
  \label{fig:objectness}
\end{figure}

Fig.~\ref{fig:objectness} complements the distribution analysis in Section~5.1 of the main paper by visualizing per-object objectness scores spatially. It shows the objectness bias induced by foreground-background supervision on COCO \texttt{val2017} images ~\cite{lin2014microsoft}. 
The first column represents the input images. The second and third columns show results from CenterNet~\cite{zhou2019objects} and FCOS~\cite{tian2019fcos}, respectively. The last column reports objectness for B-FOR trained on COCO80.
In all images, green boxes correspond to objects annotated in COCO, whereas red boxes denote objects that are missing from COCO annotations. The number within the filled circle at each object center represents the objectness score assigned by the detector. CenterNet and FCOS consistently assign low scores to unlabeled objects, indicating a strong bias caused by treating unlabeled regions as background. In contrast, B-FOR produces comparable objectness responses for both labeled and unlabeled objects, suggesting that removing explicit background supervision over unlabeled regions substantially alleviates objectness bias and enables a more category-agnostic representation of objectness.

\section{Qualitative Results}
\begin{figure}[t!]
  \centering
\includegraphics[width=0.9\textwidth]{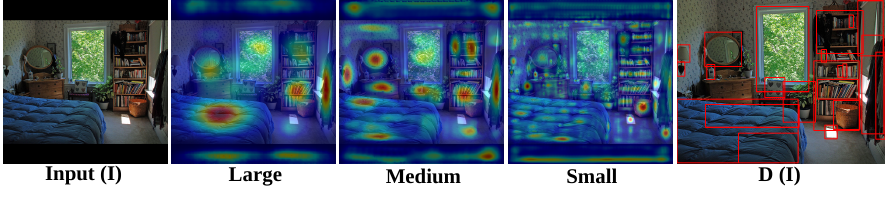}
   \caption{From left to right: input image $I$, object-center score overlays for large, medium, and small scales, and final detections $D(I)$. }
   \label{fig:inference}
\end{figure}

Fig.~\ref{fig:inference} provides qualitative evidence of the dense object-centric representations learned by B-FOR. The object-center score fields consistently activate on visually coherent object regions across multiple scales, producing dense responses that extend well beyond the sparse set of final detections. After box decoding and NMS, only a subset of these responses is retained, yielding the final detection set $D(I)$. This visualization highlights a key property of B-FOR: the underlying object-center fields encode substantially richer object evidence than what is ultimately preserved by the detection pipeline, suggesting that the model captures a broad spectrum of object hypotheses before the final selection stage.

\begin{figure}[b!]
 \centering
\includegraphics[width=0.9\textwidth]{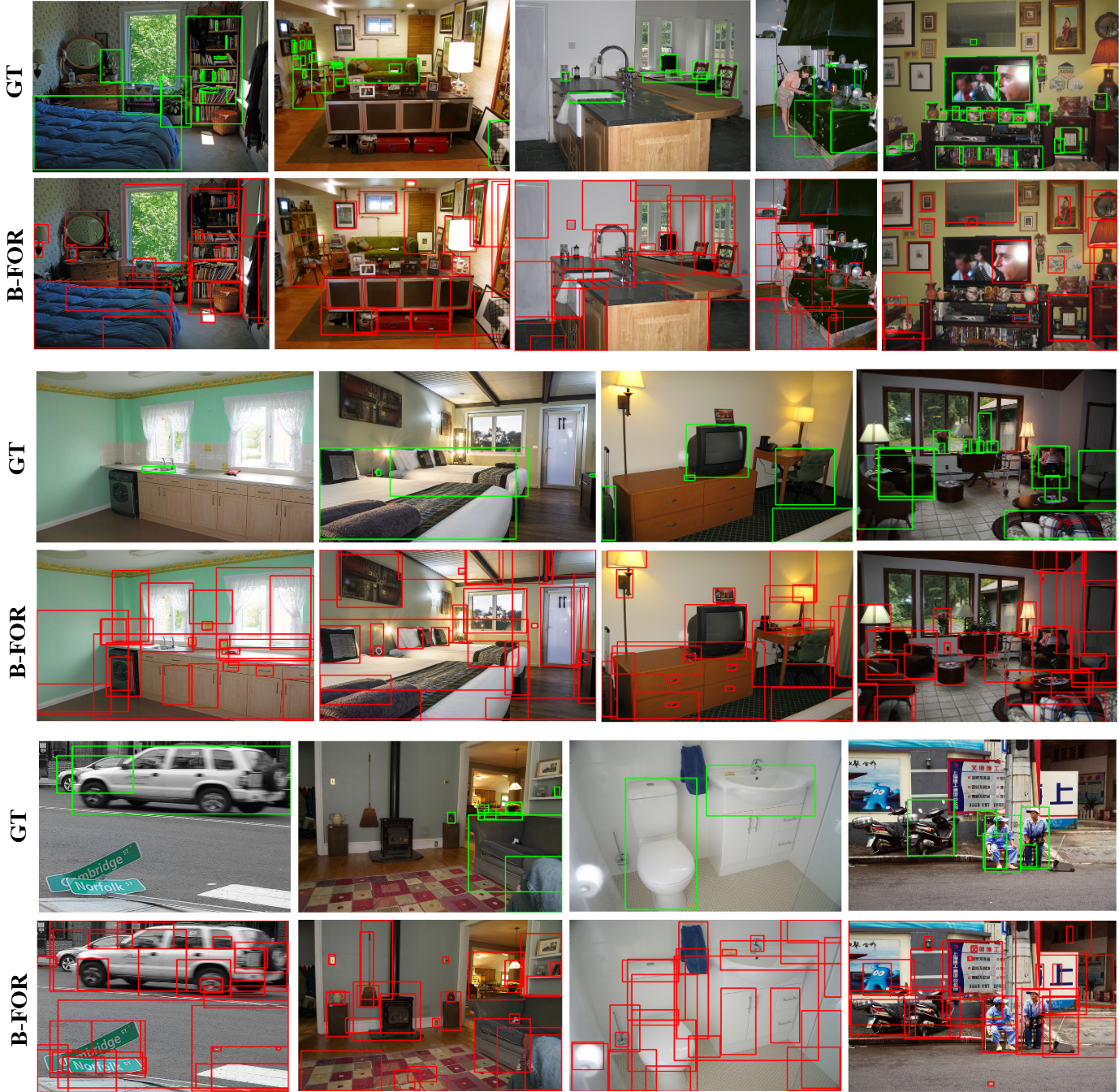}
   \caption{Qualitative top-30 detections on COCO images using a model trained on COCO80. Rows 1, 3, 5 are ground truth (GT); Rows 2, 4, 6 show the model predictions.}
  \label{fig:top30}
\end{figure}

Fig.~\ref{fig:top30} provides additional qualitative results on COCO~\cite{lin2014microsoft} validation images obtained with a model trained on COCO80. Only the top-30 detections per image are shown for readability. Ground-truth boxes are shown in green and B-FOR detections in red.  Beyond localizing annotated instances, B-FOR responds to visually coherent yet unlabeled objects.
For instance, the first image in the second row depicts a bedroom scene. The model successfully identifies several objects, including the oval mirror, wall appliques, a small bottle placed on the dresser, the heart-pattern pillow, and the basket on the floor. In the second image, it correctly detects the suitcases/boxes on the floor, the lamp, and the framed pictures. Across all images, the model consistently recognizes rectangular objects such as framed pictures, windows, cabinet doors, and electrical outlets, while also accurately detecting smaller or irregularly shaped objects, including table lamps, mirrors, and clocks.

\begin{figure*}[t!]
 \centering
\includegraphics[width=\textwidth]{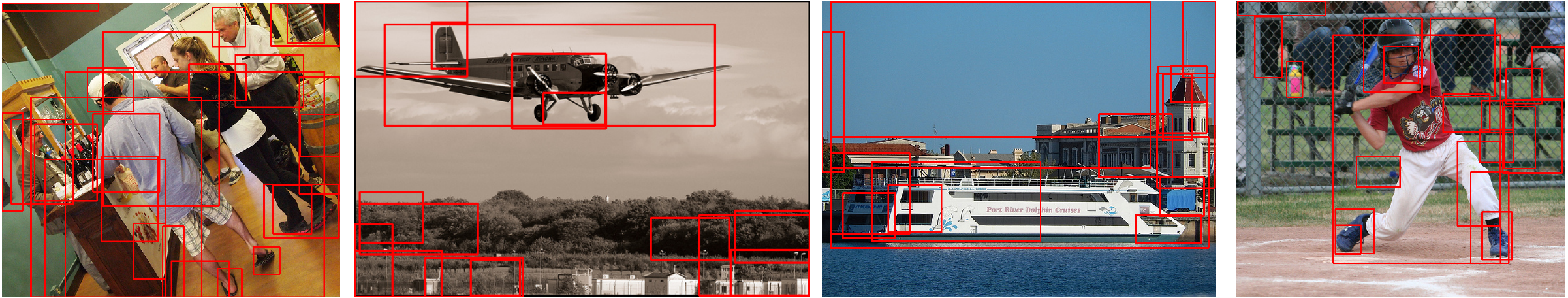}
   \caption{Failure cases: B-FOR (trained on VOC20) sometimes activates on salient object parts rather than complete instances, since box-level supervision does not distinguish objects from discriminative parts.}
  \label{fig:parts}
\end{figure*}

We also observe that the model sometimes assigns high object-center scores to salient object parts, as also shown in Fig.~\ref{fig:parts}. This occurs because B-FOR is trained from box-level object evidence without semantic class labels or part-level annotations. As a result, visually coherent and object-like parts, such as a person torso, feet, animal body region, or vehicle component, may be detected as separate object candidates. These detections are not necessarily background errors; rather, they reflect ambiguity in class-agnostic objectness, where the boundary between an object and a discriminative object part is not always explicitly defined by the supervision. However, under standard box-level evaluation, such part detections are counted as false positives if they do not sufficiently overlap a full annotated object box. This suggests that future work could improve B-FOR by incorporating object-wholeness cues or additional grouping constraints to better distinguish complete objects from their parts.

\begin{table}[b!]
\begin{center}
\scriptsize
\setlength{\tabcolsep}{4pt}
\begin{tabular}{|l|l|c|c|c|c|}
\hline
\textbf{Protocol} 
& \textbf{Training strategy} 
& \textbf{AR$_{Ovr}$} 
& \textbf{AR$_{Sml}$} 
& \textbf{AR$_{Med}$} 
& \textbf{AR$_{Lrg}$} \\
\hline\hline

\multirow{2}{*}{VOC 17 $\rightarrow$ VOC 3} 
& Staged 
& 61.5 & 43.1 & 70.2 & 60.0 \\
& Joint 
& 80.3 & 36.2 & 81.7 & 93.4 \\
\hline

\multirow{2}{*}{VOC 20 $\rightarrow$ COCO 60} 
& Staged 
& 34.7 & 11.4 & 51.6 & 56.6 \\
& Joint 
& 48.7 & 11.0 & 71.1 & 91.8 \\
\hline

\multirow{2}{*}{COCO 80 $\rightarrow$ OpenImages 490} 
& Staged 
& 30.8 & 7.9 & 43.8 & 30.2 \\
& Joint 
& 60.2 & 7.2 & 45.6 & 77.5 \\
\hline

\end{tabular}
\end{center}
\caption{
Comparison of staged and joint training across cross-dataset protocols. Staged training first optimizes object-center learning and then trains the scale-field branch, while joint training optimizes the full objective end-to-end. All settings use the same architecture and inference pipeline. We report AR@1000 at IoU $\geq 0.5$ overall and by object size.
}
\label{tab:staged_vs_joint}
\end{table}

\section{Joint vs Staged Training}
\label{sec:training_strategy}
To analyze the interaction between object-center and scale-field learning, we compare two training strategies in Table~\ref{tab:staged_vs_joint}. In staged training, the object-center branch is optimized first, while the scale-field branch is trained afterward using the learned object-center responses. This decoupled optimization separates center localization from box-extent prediction and often produces sharper intermediate score fields.
In joint training, both branches are optimized simultaneously under the full objective, enabling the scale fields to adapt directly to the object-center peaks used during inference. As a result, object-center and scale-field learning mutually reinforce each other, jointly defining objects through both their center locations and spatial extents.

Table~\ref{tab:staged_vs_joint} compares staged and joint training to isolate the contribution of simultaneous optimization. Joint training
consistently outperforms staged training across all protocols, clarifying that the final performance improvements arise from the combined objectness and scale-field supervision rather than from a particular training schedule.
\begin{figure}[b!]
  \centering
  \includegraphics[width=0.75\linewidth]{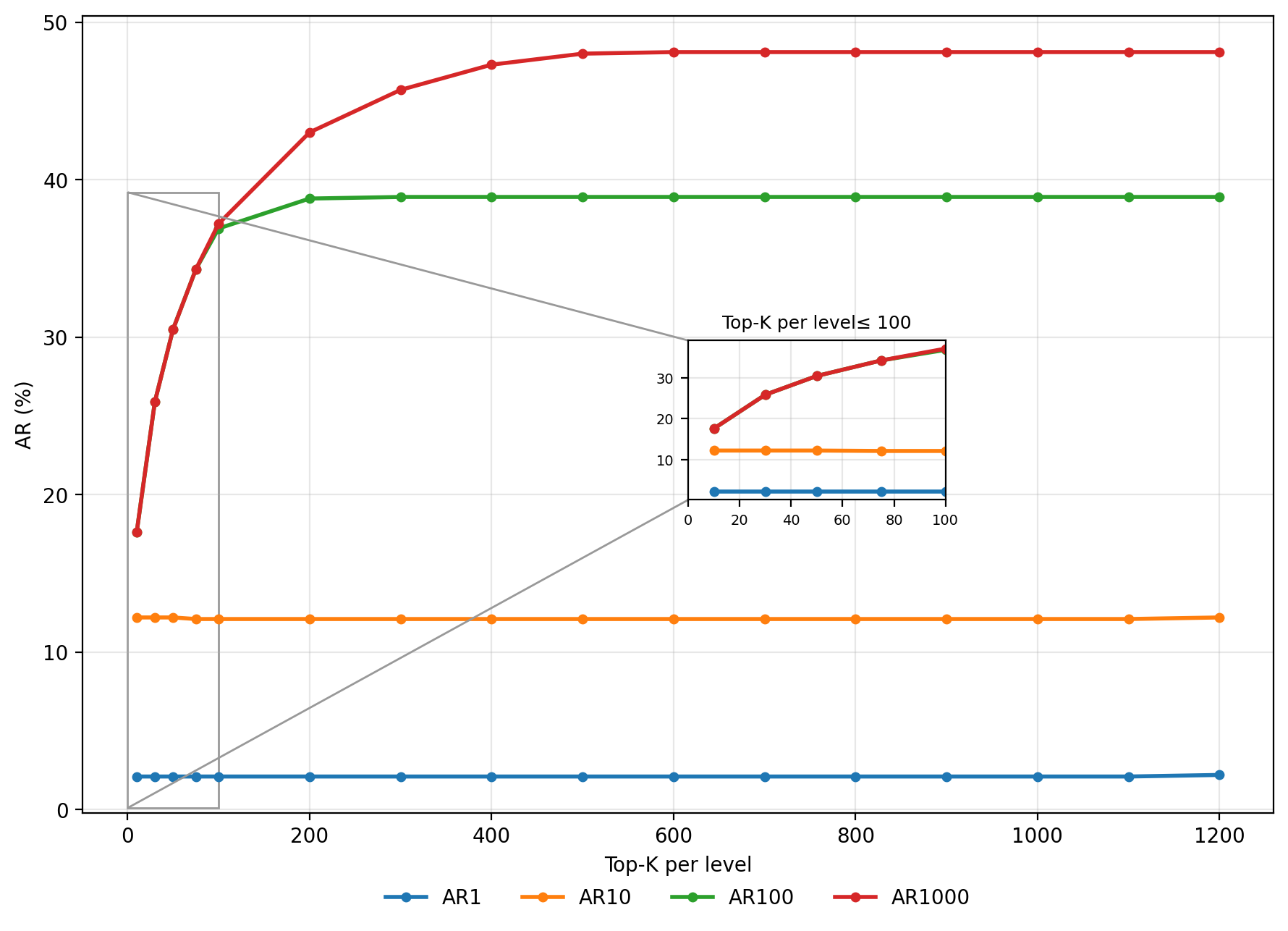}
  \caption{AR1/AR10/AR100/AR1000 vs.\ top-$K$ peaks retained per pyramid level, VOC20$\to$COCO60. All curves saturate by $K{\approx}600$; the inset zooms into $K\leq100$, where AR100/AR1000 are still rising.}
  \label{fig:topk}
\end{figure}

\section{Top-\textit{K} Analysis}
To verify that the reported recall gains do not arise from a large decoding-time candidate pool rather than an improved objectness field, we conduct a sensitivity analysis over the number of retained peaks $K$ per pyramid level. B-FOR retains the top-$K$ score peaks at each of the three levels before non-maximum suppression; if recall were driven mainly by pool size rather than objectness quality, AR would continue increasing well beyond the $K=1200$ used throughout the paper. We sweep $K \in \{10, 30, 50, 75, 100, 200, 300, \ldots, 1200\}$ on the VOC20$\to$COCO60 cross-dataset benchmark and report AR@1, AR@10, AR@100, and AR@1000 at each setting (Fig.~5). AR@1 and AR@10 are effectively constant across the full range, AR@100 plateaus by $K{\approx}200$--$300$, and AR@1000 plateaus by $K{\approx}500$--$600$; results at $K{=}1200$ are unchanged from $K{=}600$ across all four metrics. This confirms that the reported recall gains saturate well within the evaluated budget and are not an artifact of an arbitrarily large candidate pool.

\section{Objectness Score in Pseudo-Background Regions}

Reliable ground-truth background labels are unavailable under incomplete annotation; therefore, we measured objectness on conservative pseudo-background regions: COCO-Stuff locations sampled outside a safety margin from any COCO or LVIS object. B-FOR achieves $0.42\!\pm\!0.18/0.38\!\pm\!0.17/0.29\!\pm\!0.12$, compared with FCOS ($0.24\!\pm\!0.19/0.09\!\pm\!0.15/0.04\!\pm\!0.07$) and CenterNet ($0.23\!\pm\!0.23/0.08\!\pm\!0.13/0.01\!\pm\!0.01$) for
labeled/unlabeled/pseudo-background, respectively.  Moreover, candidate peaks are selected using the
image-adaptive threshold (Sec.~3.2), making detection depend on relative local maxima rather than a fixed score threshold. Taken together, these analyses show that B-FOR reduces labeled-unlabeled objectness bias while preserving discrimination from pseudo-background, inconsistent with the flattened-field hypothesis.

\section{Loss Definitions}
Table~6 in the main paper isolates the two central design choices of B-FOR by comparing against two standard baseline losses: dense foreground-background cross-entropy supervision ($\mathcal{L}_{\mathrm{CE}}$) and direct L1 box regression ($\mathcal{L}_{\mathrm{L1}}$). We define both explicitly here.

$\mathcal{L}_{\mathrm{CE}}$ denotes the conventional dense objectness loss adopted by anchor-free, center-based detectors such as CenterNet~\cite{zhou2019objects} and FCOS~\cite{tian2019fcos}: every location $q$ in the full image grid $\Omega_0$ is supervised using the same Gaussian soft target $Y(q)=\max_n Y_n(q)$, but evaluated over $\Omega_0$ rather than restricted to $\Omega(B)=\bigcup_n\Omega(b_n)$:
\begin{equation}
\mathcal{L}_{\mathrm{CE}} = -\frac{1}{|\Omega_0|}\sum_{q\in\Omega_0}
\Big[Y(q)\log\hat{s}_\ell(q) + (1-Y(q))\log(1-\hat{s}_\ell(q))\Big].
\end{equation}
This is the key methodological contrast with our proposed $\mathcal{L}_{\mathrm{ROS}}$: $\mathcal{L}_{\mathrm{CE}}$ treats every non-annotated location as an explicit negative target, whereas $\mathcal{L}_{\mathrm{ROS}}$ excludes such locations from supervision entirely.

$\mathcal{L}_{\mathrm{L1}}$ denotes direct regression of the predicted width/height to the box's true annotated size at its true center, i.e. evaluated at $q=c_n$ with the fixed target $(w_n,h_n)$, rather than the offset-compensated target $(w_q,h_q)$ (Eq.~8 and 9) used by our displacement-aware $L_{SF}$.\\

\bibliography{egbib}

\end{document}